\pdfoutput=1

\documentclass[11pt]{article}

\usepackage[final]{acl}

\usepackage{times}
\usepackage{latexsym}
\usepackage{multirow}

\usepackage[T1]{fontenc}

\usepackage[utf8]{inputenc}

\usepackage{float}

\usepackage{microtype}

\usepackage{inconsolata}

\usepackage{graphicx}

\title{CASE: Efficient Curricular Data Pre-training for Building Assistive Psychology Expert Models}

\author{
 \textbf{Sarthak Harne\textsuperscript{*1}},
 \textbf{Monjoy Narayan Choudhury\textsuperscript{*1}},
 \textbf{Madhav Rao\textsuperscript{1}},
 \textbf{TK Srikanth\textsuperscript{1}},
\\
 \textbf{Seema Mehrotra\textsuperscript{2}},
 \textbf{Apoorva Vashisht\textsuperscript{2}},
 \textbf{Aarushi Basu\textsuperscript{3}},
 \textbf{Manjit Sodhi \textsuperscript{4}}
\\
\\
 \textsuperscript{1}IIIT Bangalore,
 \textsuperscript{2}NIMHANS,
 \textsuperscript{3}Ashoka University,
 \textsuperscript{4}IBM,
\\
 \small{
   \textbf{Correspondence:} \href{mailto:sarthak.harne@iiitb.ac.in}{sarthak.harne@iiitb.ac.in}, \href{mailto:monjoy.choudhury@iiitb.ac.in}{monjoy.choudhury@iiitb.ac.in}
 }
 \\
 \small{
    \textsuperscript{*}Equal Contribution.
 }
}

\begin{document}
\newcommand{\insertfig}{\includegraphics[width=\linewidth]{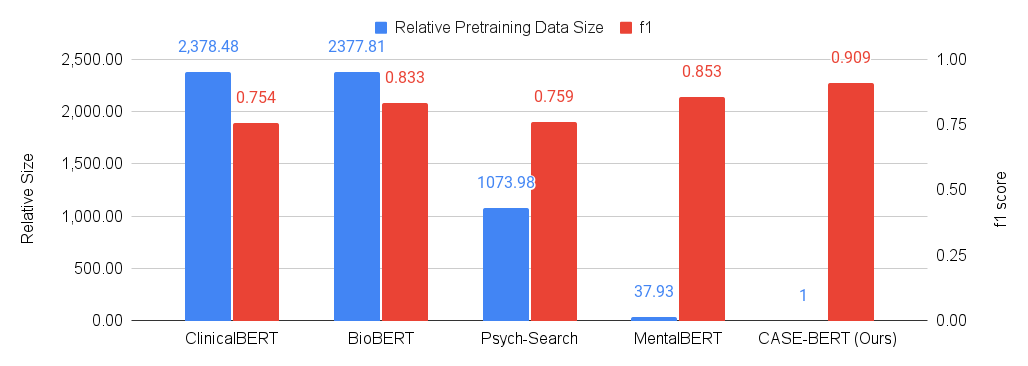}\captionof{figure}{Comparison of our pre-training data size and our model performance on the task of identifying Depression on the CounselChat dataset against ClinicalBERT \cite{clinicalbert}, BioBERT \cite{biobert}, Psych-Search \cite{Psych-Search} and MentalBERT \cite{mentalbert}. We keep our dataset size as the reference and compare the size of other datasets (based on the number of words) to it.}\label{fig:relative-data-size}}

\makeatletter
\apptocmd{\@maketitle}{\centering\insertfig}{}{}
\makeatother
\maketitle

\begin{abstract}
The limited availability of psychologists necessitates efficient identification of individuals requiring urgent mental healthcare. This study explores the use of Natural Language Processing (NLP) pipelines to analyze text data from online mental health forums used for consultations. By analyzing forum posts, these pipelines can flag users who may require immediate professional attention. A crucial challenge in this domain is data privacy and scarcity. To address this, we propose utilizing readily available curricular texts used in institutes specializing in mental health for pre-training the NLP pipelines. This helps us mimic the training process of a psychologist. Our work presents CASE-BERT that flags potential mental health disorders based on forum text. CASE-BERT demonstrates superior performance compared to existing methods, achieving an f1 score of 0.91 for Depression and 0.88 for Anxiety, two of the most commonly reported mental health disorders. Our code and data are publicly available. \footnote{https://github.com/sarthakharne/CASE}
\end{abstract}

\section{Introduction}
Mental health plays a vital role in the lifestyle of individuals worldwide. As living situations become complex, individuals around the world have expressed signs of disorders more frequently than before. The demography of mental health disorders affects every age group ranging from elementary school kids to senior citizens. In India, approximately 10.6\% of the Indian population is estimated to have mental disorders \cite{nimhansreport}. There is a huge imbalance in the number of psychologists present to address the issues in comparison to the potential number of patients seeking help. Garg et. al \cite{indiapsych} state that for every 100,000 population, there are only 0.75 professionals available in India while the required number is 3. There is a similar world statistic with only 0.5 professionals present for 100,000 population while the desired value is 2. This imbalance is further worsened as such professionals are not present in every location possible and at all times. 

\par To address the above issues, many mental health discussion forums emerged in recent times where users can express their issues and have a discussion. This can be in the form of a closed chat with humans on the other end or a chatbot along with a public forum with anonymized public posts on the front end but tracked by the platform in the backend. CounselChat\footnote{https://counselchat.com/}, Mind.org\footnote{https://www.mind.org.uk/}, and Manochikitsa\footnote{https://manochikitsa.com/} are some of the platforms that provide such facilities to users. However, the emergence of such a platform brings in the challenge of disorder identification as there is no restriction to the type of text that can be sent. Screening such texts in public forums becomes a challenging task, especially in a manual setting introducing the possibility of human error where one might miss a serious post. An automated filtration strategy can help where the suspected posts can be presented to the human in the loop for further scrutiny of the filtered posts. 

\par In this work, we aim to present a discriminative model that flags potential mental health disorders based on forum text. This work intends not to replace a mental health professional but to be used as an assistant for the preliminary screening of patients, especially in a public forum setting where previously a manual screening may have been used. However, this manual screening is costly, both in terms of human labor and time, which in the case of detecting severe cases might not be present for screening thousands of public mental health forums. In summary, our work contributes in the following ways: 

\begin{enumerate}
    \itemsep1pt
    \item We propose a domain-agnostic pre-training paradigm for discriminative models that uses curricular data from the domain to create a domain expert with a small amount of data. This domain expert model can then be fine-tuned for tasks in that domain.
    \item We compare the size of our pre-training dataset with that of other models. Fig.\ref{fig:relative-data-size} and Table \ref{tab:relative-sizes} show the size of the other pre-training dataset (with respect to the number of words) relative to ours. 
    \item We propose CASE-BERT, a BERT \cite{BERT} based mental health expert model made using curricular data pre-training which achieves SOTA performance on significantly less data as shown in Table \ref{Dep_red_table} and Table \ref{CounselChatPerformance}. CASE-BERT-Base and CASE-BERT-Small models will be made publicly available.
\end{enumerate}

\par Our work builds on the philosophy of how a person gains curricular knowledge. Our approach is analogous to the training of a psychology student as mandated by the American Psychology Association (APA) \cite{teachingclinicalpsychology} from a theoretical standpoint. Building on Textbooks Are All You Need \cite{textbooks} where they use high-quality data for building a small, yet effective generative model for Python Programming, we pre-train off-the-shelf models using high-quality curricular data provided by psychologists in academia. This high-quality data helps us to tackle domain data scarcity by using a small amount of data to pre-train and finetune state-of-the-art (SOTA) discriminative assistive psychology expert models. We believe our contribution in the form of curriculum learning for discriminative models can be used in other domains also, given the amount of curriculum data present in the domain is respectable.

\par We compare our work on several mental health detection datasets with several other similar works like MentalBERT \cite{mentalbert}, Psych-Search \cite{Psych-Search}, BioBERT \cite{biobert} and ClinicalBERT \cite{clinicalbert}.

\begin{figure*}[t]
    \centering
    \includegraphics[width=0.85\linewidth]{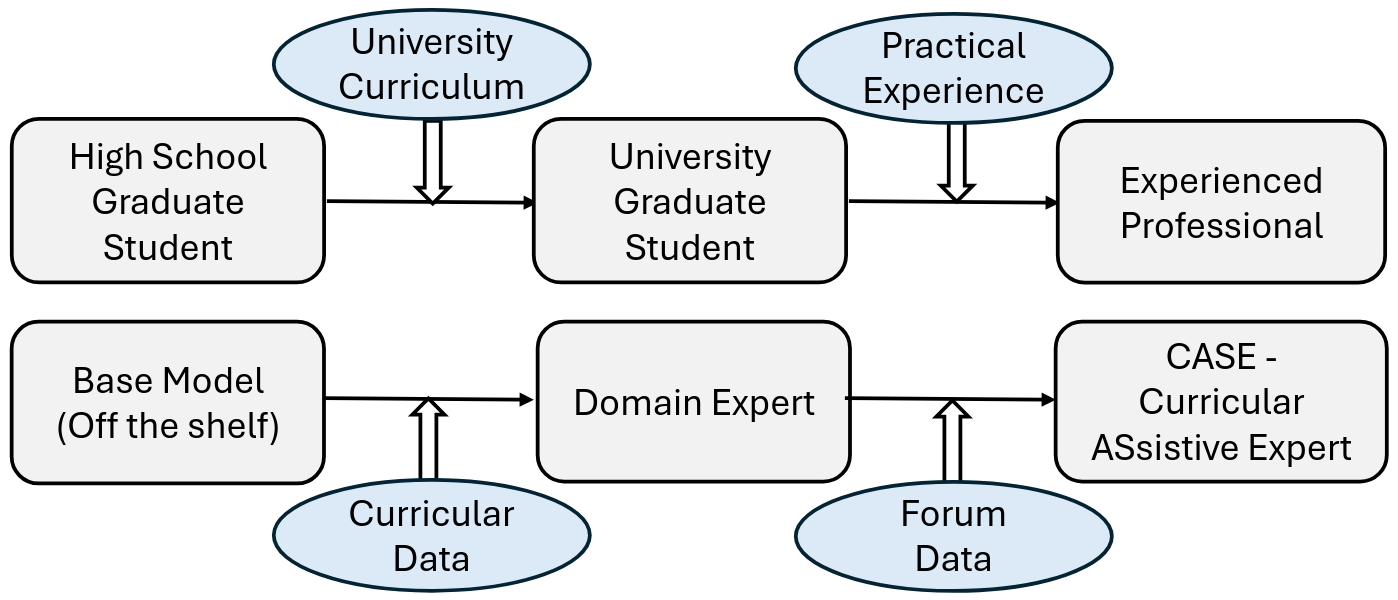}
    \caption{The analogy between our pre-training philosophy and the steps of training required to become an experienced professional. In contrast to previous work, we ensure that relevant data is provided to the model to identify and understand the task during the pre-training stage.}
    \label{fig:pretraining-method}
\end{figure*}

\par The structure of the rest of this article is as follows: Section \ref{Section2} presents the background about the baseline models, and related literature and methods used in our approach. Section \ref{section3} discusses the dataset collection process used for pre-training and fine-tuning all model instances discussed. Section \ref{section4} discusses our approach for CASE-BERT. Section \ref{section5} presents the set of experiments performed and evaluation metrics used to support our hypothesis. Lastly, Section \ref{section6} discusses the results obtained along with ethical concerns and presents a line of future work associated with our work.

\section{Background and Related Work}
\label{Section2}
\par Prior work uses data sources like mental health subreddits \cite{mentalbert}, abstracts of scientific articles on psychiatry \cite{Psych-Search}, \cite{biobert} and full texts of scientific articles on psychiatry \cite{biobert}.

\subsection{Discriminative Models for Biomedical applications}
Prior work has introduced several publicly available models geared for Biomedical applications. These range from text mining \cite{biobert}, natural language inference \cite{clinicalbert}, natural language understanding \cite{Psych-Search} and mental health disorder identification \cite{mentalbert}.

\subsubsection{Discriminative Clinical and Biomedical Models}
BioBERT \cite{biobert} uses PubMed Central articles and PubMed article abstracts for pre-training to create a model for text mining for practical use in biomedical applications like NER, Relation Extraction and Question Answering. ClinicalBERT \cite{clinicalbert} uses notes from patient discharge summaries to further pre-train BioBert \cite{biobert} for MedNLI tasks.

\subsubsection{Discriminative Mental Health Models}
MentalBERT \cite{mentalbert} leverages a pre-trained BERT model (bert-base-uncased\footnote{https://huggingface.co/google-bert/bert-base-uncased}) and fine-tune it on a dataset of 13 million sentences scraped from Reddit \cite{reddit} communities specifically focused on mental health issues like depression, anxiety, and bipolar disorder. Psych-Search \cite{Psych-Search} an open-sourced model is based on SciBERT \cite{scibert}, a model pre-trained on a collection of scientific paper abstracts. Psych-Search refines SciBERT further by specifically training it on 3.2 million abstracts related to psychology and psychiatry obtained from PubMed. 
\par To the best of our knowledge, no previous work has been done where the pre-training stage uses curricular data and studied the effects of the same. Just using Reddit posts and PubMed abstracts might not necessarily give the model knowledge about Psychology. Reddit posts merely contain a person talking about their problems and article abstracts from PubMed are too short to contain a lot of useful information and instead focus on summarizing the contents of the article.

\subsection{Textbooks Are All You Need}
Textbooks Are All You Need \cite{textbooks} introduces the notion of how the quality of data affects the performance of text generation of Large Language Models. In this work, they rely highly on the hypothesis that involves the usage of "textbook quality" data from the web (6B tokens) as well as synthetically generated data from GPT3.5. This small-scale data compared to large web-crawled corpora gave a respectable performance, especially for a significantly lower number of parameters. They pre-train on such data and then attempt to fine-tune on "textbook-exercise" like data. In this way they attempt to capture the domain learning process for humans. We aim to verify this ideology in the case of mental health disorder identification in discriminative models.

\section{Data}  
\label{section3}
\begin{table*}[!htbp]
\centering
\resizebox{\textwidth}{!}{%
\begin{tabular}{c|c|c|c|c}
\hline
\textbf{Models}           & \textbf{\#Words (Approx)} & \textbf{\#Sentences (Approx)} & \textbf{Relative Size} & \textbf{Depression(f1)} \\ \hline 
ClinicalBERT \cite{clinicalbert}             & 18.10 Billion                        & 873 Million                           & 2378.48x               & 0.754                    \\ 
BioBERT \cite{biobert}                  & 18.15 Billion                        & 857 Million                           & 2377.81x               & 0.833                    \\ 
Psych-Search \cite{Psych-Search}             & 8.13 Billion                      & 387 Million                           & 1073.98x               & 0.759                    \\ 
MentalBERT \cite{mentalbert}               & 2.87 Billion                      & 13.7 Million                          & 37.93x                 & 0.853                    \\
\textbf{CASE-BERT (Ours)} & \textbf{7.57 Million}             & \textbf{0.36 Million}                 & \textbf{1x}            & \textbf{0.909}           \\ \hline 
\end{tabular}
}
\caption{Comparison of our pre-training data size (lower being better) and our model performance on the task of identifying Depression on the CounselChat dataset. The number of words and sentences is an approximation to the nearest ten thousand.}
\label{tab:relative-sizes}
\end{table*}
Due to the sensitive nature of the data required for building models for mental health disorder diagnosis, relevant good quality counseling data (like transcripts of sessions between therapists and patients) is difficult to find and to the best of our knowledge, not available publicly. Moreover, as protecting the patient's confidentiality is paramount, using such data is unethical unless explicit consent is received from all the involved parties. 

\par Creating a counseling dataset is difficult and costly as discussed above. Hence, we limit our use case to screening patients on mental health forums or social media as publicly available data of this form is available. The dataset used for this work involves publicly available datasets and data curated by psychology professionals.
\par Mental Health is a sensitive domain with a severe lack of open-source well-annotated data and unsupervised datasets that can be used as corpora for pre-training. We hypothesise that our philosophy of using curricular data effectively creates domain experts. We test this hypothesis specifically for the domain of Mental Health disorder detection. Future work can involve testing this hypothesis on other domains that suffer from scarcity of data due to its sensitive nature. We discuss the datasets used in detail below. 

\subsection{Curricular Dataset}
\label{curricular-dataset}
We created a private curricular text dataset with the help of the clinical psychologists who collaborated in this work. The psychologists were a combination of people working in academia as well as professional psychiatric practitioners. We collected a set of text materials with the advice of these psychologists which met certain criteria explained subsequently.

\par The dataset is created in such a manner that it contains text that covers a wide area of the study in psychology. This dataset consisted of 110 curricular text materials that are used to train Psychology students in graduate-level education. The books focus on general knowledge of psychology and cover a wide range of topics. As the model's intended use is to identify signs of mental health disorders, we also include a number of books on interviewing and counseling skills in psychotherapy. For integrating more nuanced knowledge, we also include several journal articles, awareness pamphlets and brochures and published seminal papers on various topics. We cover the following topics: 1. Psychology, 2. Interviewing and Counselling, 3. Cognition and Computation, 4. Clinical Child and Adolescent Psychology, 5. Clinical Psychology, 6. Psychotherapy, 7. Telepsychotherapy, 8. Society, Collective Sociology, and Psychology, 9. Geriatric Counselling. We understand that the outlook towards Mental Health is different across different countries, hence we include texts from various parts of the world, specifically North America, Asia, and Europe. The curriculum across different countries is very identical to each other. We follow the guidelines of the American Psychology Association's \cite{teachingclinicalpsychology} recommendation due to their clear and in-depth documentation of their recommended curriculum and teaching methods.



\par One of the aims of this work is to present a pre-training philosophy that can be applied to domains that suffer from data scarcity. As a result, we believe it is important that the method is not data-hungry even during the pre-training stage. The pre-training dataset has 7,567,108 words or 365,937 sentences. We compare the size of our pre-training dataset with that of other models. Fig.\ref{fig:relative-data-size} and Table.\ref{tab:relative-sizes} show the size of the other pre-training dataset (with respect to the number of words) relative to ours. 
This data efficiency can be crucial to many domains and is certainly useful in the psychology domain.

\subsection{CounselChat Dataset}
CounselChat is an expert community platform. It serves as a platform to assist counselors in establishing connections with possible clients. Therapists answer queries from customers on the website. It's a good concept that produces some fascinating data.
\par A lot of prior work done previously uses data from Reddit to test their model's ability to identify mental health disorders \cite{mentalbert}, \cite{Psych-Search}, \cite{twostreamdetection}. Some of these datasets \cite{swmh}, \cite{depression-reddit} use the post's subreddit to indicate whether signs of the mental health disorder were present or not. This assumption is essential to automate the data collection process which gives a dataset of considerable size. However, this assumption does lead to many incorrect labels. On the other hand, CounselChat is a forum specifically for discussing one's problems. The tags given to these more closely reflect the state of a person's mental health disorder.
\par The posts from CounselChat have been scraped and made public by Bertagnolli \cite{counselchat}. We use the query text here for our model fine-tuning along with the tags for the discriminative model. It consists of 1374 rows with an average passage size of 140 words. These sentences are tagged with multiple labels ranging from depression, anxiety, addiction, marriage, relationships, and so on. Out of these, we aim to use the data points tagged depression and anxiety as the classes for our model, as these are the most common disorders in the world by a large margin as reported by WHO \cite{anxietydepressionprevalencewho} and Institute of Health Metrics and Evaluation \cite{anxietydepressionprevalenceGHDx}. This involves extracting all texts marked with the above-mentioned tags as positive examples while the other samples that lack the tag are considered to be negative examples resulting in 198 and 231 rows labeled true for Depression and Anxiety respectively in this dataset. 

\subsection{Other Datasets on Depression and Stress}
Previous work \cite{mentalbert} has used datasets like Depression\_Reddit \cite{depression-reddit} and Dreaddit \cite{dreaddit} to fine-tune and test their models. Depression\_Reddit \cite{depression-reddit} data focuses on identifying signs of depression from Reddit posts. Dreaddit \cite{dreaddit} on the other hand focuses on identifying signs of stress from Reddit posts from five different forums. We report the performance of our model on these datasets for consistency with the prior work.

\section{Method}
\label{section4}
As mandated by APA \cite{teachingclinicalpsychology} clinical psychology student undergoes years of training where they refer to various curricular documents to learn the mandated clinical psychology theory, like textbooks, as prescribed by their institution. This is then followed by supervised clinical experience in handling patients as subordinates, in the form of fellowships where they work to diagnose several (at least 100) patients; and finally becoming practitioners.
\par Our approach tries to mimic this. We consider the off-the-shelf BERT \cite{BERT} model similar to a student graduating from high school, having a general knowledge of the world. Next, we use curricular data mentioned in Section \ref{curricular-dataset}. After this pre-training, we consider our model to be a psychology student having theoretical knowledge of psychology and related fields like psychotherapy. However, it does not have practical experience with patients. For this purpose, we further fine-tune it on the CounselChat \cite{counselchat}, Dreaddit \cite{dreaddit} and Depression\_Reddit \cite{depression-reddit}.

\subsection{Pre-training}
We build upon previous work \cite{textbooks} on pre-training using textbook-like data generated using GPT-3.5\footnote{https://openai.com/index/gpt-3-5-turbo-fine-tuning-and-api-updates/} for building a generative model (decoder only) for Python Programming. Our work takes a similar approach but diverges at the point where we try to tackle domain data scarcity by using curricular data for building a discriminative model. We use the same technique of mask language modeling as BERT \cite{BERT} to pre-train our model.

\subsection{Fine-tuning}
A simple Multi-Layer Perceptron with 2 layers was used as a sequence classification head for the binary classification problem of detecting whether certain posts exhibit the possibility of the above-mentioned disorders. The datasets we use for fine-tuning involve the task of binary classification -- detecting whether a mental health disorder is present or not.
\begin{figure}[h]
    \includegraphics[width=\linewidth]{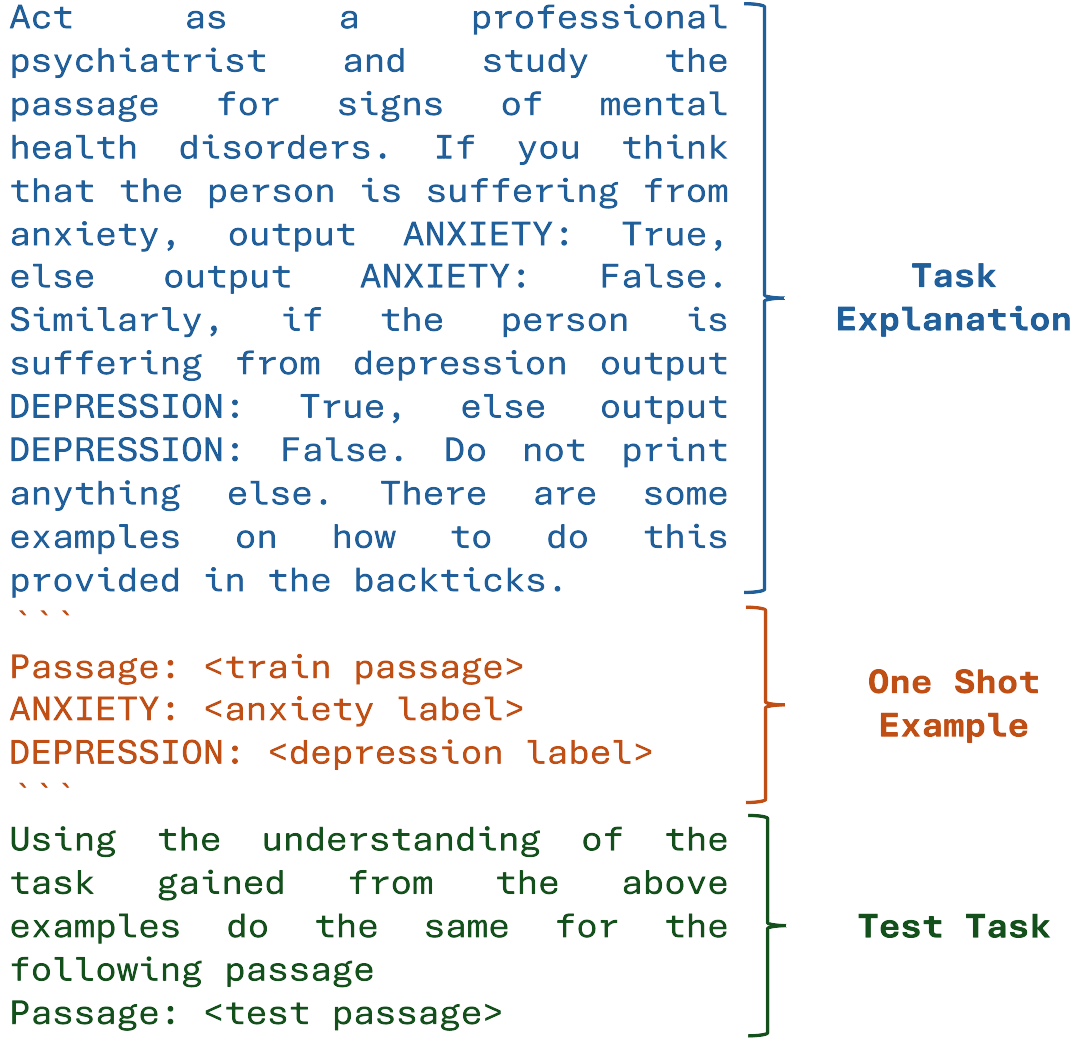}
    \caption{The prompt template used for inference on CounselChat. Relevant changes are made in the prompt to be compatible with the other datasets viz. the disorder/symptoms to be identified.}
    \label{fig:few-shot-prompt}
\end{figure}
\subsection{Prompting}
Due to the recent availability of capable instruction-tuned LLMs, \cite{gpt35}, \cite{llama38b} it has become easier to solve general problems via merely prompting these models \cite{chatgptsurvey}. To compare how well these discriminative models perform against modern generative LLMs, we test these LLMs on the same datasets. For this, we use a few-shot prompting strategy. The prompt structure used for inference on CounselChat \cite{counselchat} is shown in Fig.\ref{fig:few-shot-prompt}. We first experimented with various prompt styles and picked this prompt based on some subjective analysis. We choose a random instance from the train split of the dataset to be passed on as the one-shot example. Subsequently, we pass one instance from the test set for the model to infer.







\section{Experiments}
\label{section5}

\subsection{Pre-training}
\par We employ the mask language modeling task as discussed in BERT \cite{BERT} on the curricular data where we masked tokens randomly with a probability of 0.15. We take an off-the-shelf BERT model \cite{BERT} (bert-base-uncased) and use the curricular dataset for pre-training. We first break the data into windows of 500 tokens with an overlap of 50 tokens between subsequent windows. We use a Nvidia P100 workbench for 60 epochs which took about 8.3 hours to train. Gradient accumulation is applied to achieve an effective batch size of 128. Learning rate of $1\times10^{-5}$ was used. 

\par To test the utility of our pre-training, we compare the results of fine-tuning on an off-the-shelf BERT \cite{BERT} model. We pre-train both the bert-base-uncased and the bert-small-uncased model.

\subsection{Fine-tuning}
To test the performance of our model we test it on CounselChat \cite{counselchat}, Depression\_Reddit \cite{depression-reddit}, and Dreaddit \cite{dreaddit} after fine-tuning it on train splits of these datasets. We train this on a workbench with two Nvidia T4 GPUs for 3 epochs with a batch size of 32. The learning rate used was the same as that of the pretraining step.

\par We compare our results against an off-the-shelf BERT \cite{BERT} model, off-the-shelf RoBERTa model \cite{roberta}, MentalBERT \cite{mentalbert}, Psych-Search \cite{Psych-Search}, BioBERT \cite{biobert} and ClinicalBERT \cite{clinicalbert}. We finetune these models for using the above datasets with the same hyperparameters to ensure uniformity in comparison. We report the metrics on the same test split on all the models. We call the fine-tuned models as CASE-BERT-Base and CASE-BERT-Small. Comparing CASE-BERT-Base shows us the advantage of using our pre-training philosophy. We intend to compare CASE-BERT with MentalBERT \cite{mentalbert} and PsychSearch \cite{Psych-Search} as they are tuned specifically for tasks in psychology. We compare with BioBERT \cite{biobert} and ClinicalBERT \cite{clinicalbert} to see how models meant for general biomedical use perform in the domain of Psychology.



\subsection{Prompting}
We use the prompt template shown in Fig. \ref{fig:few-shot-prompt} for running inference on CounselChat \cite{counselchat} dataset. Depression\_Reddit \cite{depression-reddit} contains the labels for depression only. Similarly, Dreaddit \cite{dreaddit} contains a label for stress only. Appropriate changes are made to the prompt so that it is made suitable for these datasets. We choose GPT-3.5 \cite{gpt35}, Llama 3 8b Instruct \cite{llama38b}, Mistral \cite{mistral} and Gemma 7b Instruct \cite{gemma} to compare our models against.

\subsection{Evaluation}
We use the f1-score, precision, recall, and accuracy to evaluate our model against the baselines previously mentioned.

\begin{table*}[t]
\resizebox{\textwidth}{!}{%
\begin{tabular}{cc|cccc|cccc}
\hline
\multicolumn{1}{l}{\multirow{2}{*}{}} & \multicolumn{1}{l|}{}           & \multicolumn{4}{c|}{\textbf{Depression\_Reddit}}                          & \multicolumn{4}{c}{\textbf{Dreaddit}}                                         \\ \cline{3-10} 
\multicolumn{1}{l}{}                  & \textbf{Model Name}            & \textbf{Recall} & \textbf{Precision} & \textbf{f1 score} & \textbf{Accuracy} & \textbf{Recall} & \textbf{Precision} & \textbf{f1 score} & \textbf{Accuracy} \\ \hline
\multirow{4}{*}{\rotatebox{90}{Generative}}           & Mistral 7b                     & 0.228           & 0.407             & 0.292             & 0.225             & 0.293           & 0.312             & 0.302             & 0.302             \\ 
                                      & Llama 3 8b                     & 0.733           & 0.585             & 0.650             & 0.611             & 0.813           & 0.550             & 0.656             & 0.561             \\ 
                                      & Gemma 7b instruct              & 0.660           & 0.886             & 0.757             & 0.702             & 0.675           & 0.570             & 0.618             & 0.569             \\ 
                                      & GPT-3.5-Turbo-1106             & 0.830           & 0.896             & 0.862             & 0.813             & \underline{0.986}           & 0.609             & 0.753             & 0.666             \\ \hline
\multirow{8}{*}{\rotatebox{90}{Discriminative}}       & BERT                           & 0.911           & 0.907             & 0.909             & \textbf{0.951}             & 0.818           & 0.786             & 0.802             & 0.792             \\ 
                                      & RoBERTa                        & 0.951           & \underline{0.952}             & 0.951             & \underline{0.949}             & 0.843           & 0.787             & \underline{0.814}             & \underline{0.801}             \\ 
                                      & Mental-BERT                    & 0.946           & 0.946             & 0.946             & \textbf{0.951}    & 0.789           & \textbf{0.808}    & 0.798             & 0.794             \\ 
                                      & Psych-Search                   & 0.950           & 0.946             & 0.948             & 0.927             & \textbf{1.000}  & 0.516             & 0.681             & 0.516             \\ 
                                      & BioBERT                        & \underline{0.961}           & 0.943             & \underline{0.952}             & 0.932             & 0.799           & 0.751             & 0.774             & 0.759             \\ 
                                      & ClinicalBERT                   & 0.958           & 0.939             & 0.948             & 0.927             & 0.851           & 0.717             & 0.778             & 0.750             \\ \cline{2-10}
                                      & CASE-BERT-Small (ours)         & 0.958           & 0.929             & 0.943             & 0.919             & 0.843           & 0.740             & 0.788             & 0.766             \\ 
                                      & \textbf{CASE-BERT-Base (ours)} & \textbf{0.969}  & \textbf{0.962}    & \textbf{0.965}    & \textbf{0.951}    & 0.846           & \underline{0.790}             & \textbf{0.817}    & \textbf{0.804}    \\ \hline
\end{tabular}%
}
\caption{Performance Comparison using Recall, Precision, f1 score and Accuracy on the two datasets Depression\_Reddit \cite{depression-reddit} and Dreaddit \cite{dreaddit}. The best-performing numbers are highlighted in bold while the second best are underlined.}
\label{Dep_red_table}
\end{table*}

\section{Results and Discussion}
\label{section6}

As the datasets used suffer from a large class imbalance, we focus on the f1 score and recall. Studying recall in such a use case is important as we need to avoid false negatives and miss people who suffer from a disorder. Having said that, we also need to maintain a high value for precision and avoid false positives so that we minimise the stress and stigma that might, unfortunately, be associated with a positive mental health disorder diagnosis.
\par The main aim of this work is to introduce our pre-training philosophy and verify the same for the use case of mental health disorder detection. Note that our aim is not to achieve the best state-of-the-art numbers on these datasets, but we still manage to have an advantage in terms of model performance over previous work.
\par Table \ref{fig:relative-data-size} and Fig.\ref{fig:relative-data-size} show the data efficiency of our method. With just a small fraction of the pre-training dataset size, we are able to achieve competitive performance as compared to previous work done in this field. This supports our hypothesis of using curricular data for pre-training -- even a small amount of data results in a much better performance. This data efficiency can be essential in creating domain experts for very niche and nuanced areas.
\par Table \ref{Dep_red_table} shows the results of our experiments on Dreaddit \cite{dreaddit} and Depression\_Reddit \cite{depression-reddit}. We see that CASE-BERT-Base outperforms previous work in terms of the f1 score and maintains similar competitive performance across precision and recall as well.
\par Table \ref{CounselChatPerformance} shows the results of our experiments on CounselChat. Just like Dreaddit \cite{dreaddit} and Depression\_Reddit \cite{depression-reddit} we see that CASE-BERT-Base maintains a lead in terms of the f1 score and performs much better as compared to the previous work. Interestingly, we also see that CASE-BERT-Small outperforms most of the other baselines even after having about half as many parameters as the other models.
\par In all of the experiments we observe that the generative models fail to perform competitively. Specifically, we can see that all the models have a higher recall than precision. This indicates that these models are more sensitive towards signs of Mental Health and are biased towards flagging the users as having a Mental Health disorder. An argument can be made that the discriminative models are fine-tuned specifically for these datasets leading to their improved performances, however, fine-tuning these LLMs with many billions of parameters is far more complex and compute-intensive than fine-tuning discriminative models which have little above a hundred million parameters. We can see how we could easily train the discriminative models on publicly available GPUs on Kaggle. Doing the same for the LLMs we compared against is not feasible on Kaggle.

\begin{table*}[ht]
\centering
\resizebox{\textwidth}{!}{
\begin{tabular}{cc|cccc|cccc}
\hline
\multicolumn{2}{c|}{} & \multicolumn{4}{c|}{\textbf{Depression}} & \multicolumn{4}{c}{\textbf{Anxiety}} \\ \cline{3-10} 
\multicolumn{2}{c|}{\multirow{-2}{*}{\textbf{Model Name}}} & \textbf{Recall} & \textbf{Precision} & \textbf{f1 score} & \textbf{Accuracy} & \textbf{Recall} & \textbf{Precision} & \textbf{f1 score} & \textbf{Accuracy} \\ \hline
\multirow{4}{*}{\rotatebox[origin=c]{90}{Generative}} & Mistral 7b & 0.289 & 0.643 & 0.105 & 0.322 & 0.293 & 0.068 & 0.111 & 0.301 \\ 
& Llama 8b Instruct & \textbf{0.974} & 0.272 & 0.425 & 0.638 & 0.756 & 0.143 & 0.240 & 0.290 \\ 
& Gemma 7b Instruct & 0.526 & 0.101 & 0.169 & 0.290 & 0.585 & 0.114 & 0.190 & 0.261 \\ 
& GPT-3.5-Turbo-1106 &\textbf{0.974} & 0.234 & 0.378 & 0.558 & \textbf{0.902} & 0.266 & 0.411 & 0.616 \\ \hline
\multirow{8}{*}{\rotatebox[origin=c]{90}{Discriminative}} & BERT & 0.737 & 0.966 & 0.836 & 0.960 & 0.829 & 0.872 & 0.850 & 0.957 \\ 
& RoBERTa & 0.842 & 0.742 & 0.789 & 0.942 & 0.642 & 0.857 & 0.734 & 0.924 \\ 
& Mental-BERT & 0.763 & \underline{0.967} & 0.853 & 0.964 & 0.854 & 0.860 & 0.857 & 0.957 \\ 
& Psych-Search & 0.789 & 0.732 & 0.759 & 0.931 & 0.854 & 0.874 & \underline{0.864} & 0.960 \\ 
& BioBERT & 0.789 & 0.882 & 0.833 & 0.957 & 0.829 & \underline{0.883} & 0.855 & \textbf{0.971} \\ 
& ClinicalBERT & 0.684 & 0.839 & 0.754 & 0.938 & 0.829 & 0.829 & 0.829 & 0.949 \\ \cline{2-10}
& \textbf{CASE-BERT-Small (ours)} & 0.842 & 0.914 & \underline{0.877} & \underline{0.967} & 0.829 & \textbf{0.895} & 0.861 & 0.960 \\ 
& \textbf{CASE-BERT-Base (ours)} & \underline{0.856} & \textbf{0.969} & \textbf{0.909} & \textbf{0.975} & \underline{0.887} & 0.875 & \textbf{0.881} & \underline{0.964} \\ \hline
\end{tabular}
}
\caption{Performance Comparison using Recall, Precision, f1 score and Accuracy on the Counsel Chat Dataset \cite{counselchat} for the classes Depression and Anxiety. The best-performing numbers are highlighted in bold while the second best are underlined.}
\label{CounselChatPerformance}
\end{table*}

\par \textbf{Discussion} In this work, we present a pre-training method for a discriminative model. By using curricular data for pre-training, we present models -- CASE-BERT-Base and CASE-BERT-Small, two discriminative BERT-based models that flag potential mental health disorders based on forum texts. We tackle the lack of granular data used for training classification pipelines and large corpora for pre-training discriminative pipelines by using curricular text data. This method of pre-training achieves much better performance with a tiny fraction of the data as compared to the previous SOTA methods.  Our work is a significant step towards building small expert models that need a small amount of data while not requiring a massive amount of computing power to deploy. We believe that the approach of using curricular data to battle data scarcity is a big challenge and has greater implications in building machine-learning solutions for domains where data privacy and security are critical. These are qualities that are essential for a model that is deployed on a Mental Health forum. Having a small model is also advantageous as it reduces or at best avoids the requirement of expensive hardware to run on.

\subsection{Limitations}
The quality of the outputs of our work however depends on the robustness of the underlying base models used -- BERT\cite{BERT} for the discriminative task. This brings in the concern of biases that the underlying language model brings. Particularly our model has a bias toward predicting a higher risk of self-harm tendencies. We discuss the ethical considerations and further scope of this work next. 
\par We hypothesize that this model can be used to create domain experts across many different areas. We verify this claim only in the domain of Mental Health. Verifying this claim in other domains as well is essential and useful.

\subsection{Ethical Concerns}
Our work attempts to assist in flagging forum texts. Mental health is a privacy-sensitive domain. However, CASE-BERT is fine-tuned on data obtained publicly from open domains that are created after anonymizing all personally identifiable information. We would also like to highlight that our models do not aim to replace professional psychologists but rather aim to assist a psychologist in screening potential patients faster than manually screening with the help of volunteers, interns, or students. As psychological evaluations are unique for every patient, we believe that our work is a step towards accelerating the preliminary step which enhances the efficient use of time of a psychologist by screening the patients.
\par Solving this problem is essential, but at the same time, we would argue that proper testing and avoiding both false negatives and false positives is essential to avoid unnecessary stress and stigma for the patient involved. Moreover, as we include texts from different countries to pre-train our model, we would recommend further pre-training with locally available text before deploying the model.

\subsection{Further Scope}
Our work opens up the possibility of creating preliminary screening pipelines that can be deployed as web applications. We can use variants of BERT models, like CASE-BERT, that take relatively less memory and various on-device assistive applications can be created as a consequence of this. In general, we believe curricular data can be used in other domains where high-quality data for a well-defined task is absent. One similar example can be in the domain of legal issue analysis where one can create such assistive screening pipelines and assign the severity of issues which would help in allocation of lawyers and accelerating the legal process for minor cases. 
\par We hope that this pre-training philosophy can be extended to generative models as well, which can generate diagnosis given a text, instead of just flagging a post for signs of a disorder. Lastly, we hope that our work opens up the avenue for the creation of more curricular training-based assistive experts that can leverage existing LLMs that can be trained on relatively accessible hardware and obtain respectable performance with a significantly lower amount of labelled data. 

\bibliography{custom}



\end{document}